# Localizing the conceptual difference of two scenes using deep learning for house keeping usages

Ali Atghaei
AI software researcher
Veunex company
Berlin, Germany
ali.atghaei@veunex.com

Ehsan Rahnama
AI software researcher
Veunex company
Berlin, Germany
ehsan@veunex.com

Kiavash Azimi
AI software researcher
Veunex company
Berlin, Germany
kiavash.azimi@veunex.com

*Abstract* - **Finding the conceptual difference between the two images in an industrial environment has been especially important for HSE purposes and there is still no reliable and conformable method to find the major differences to alert the related controllers. Due to the abundance and variety of objects in different environments, the use of supervised learning methods in this field is facing a major problem. Due to the sharp and even slight change in lighting conditions in the two scenes, it is not possible to naively subtract the two images in order to find these differences. The goal of this paper is to find and localize the conceptual differences of two frames of one scene but in two different times and classify the differences to addition, reduction and change in the field. In this paper, we demonstrate a comprehensive solution for this application by presenting the deep learning method and using transfer learning and structural modification of the error function, as well as a process for adding and synthesizing data. An appropriate data set was provided and labeled, and the model results were evaluated on this data set and the possibility of using it in real and industrial applications was explained.**

I. INTRODUCTION

One of the most practical tasks that can be done simply by the person in charge of monitoring an environment is to identify the conceptual change of the two scenes. For example, the observer can receive two images of a place, the first related to the morning (bright) and the second related to the afternoon (darker and noisier, etc.), and understand which objects in these two images were added or subtracted or changed And locate them, although he may not know the name and characteristics of the object [1].

This is not easy to do for computer systems and requires a high level of conceptual understanding and generalization of the model.

The difference between two scenes in crowded environments is such that by simply subtracting two images, it is almost impossible to get an acceptable pattern, because changing lighting conditions, changing the camera angle, the presence of shadows, change the two images, but conceptually this changes are not significant. For example, the appearance of an object may be different in various illumination conditions, but this has no effect on the presence or absence of that object in those scenes.

In this article, we are looking for a way to detect and locate objects that have been added or removed by receiving two crowded images from different times of the two scenes.

With the advancement of machine learning and deep learning technology, many problems are solved by supervised learning. However, in the mentioned application, which is mainly used in industrial centers for spatial care of the project environment, due to the excessive number and unpredictability of objects in the environment [2], we can not simply label different objects and look for them in the picture. Therefore, we are looking for a model that, regardless of identifying objects, detects their removal or addition in two different scenes.

For example, in an industrial environment, a scenario might be defined in such a way that if the work environment is not cleaned or sorted at the end of the day, or some objects are added or removed, the technical and security expert of the workshop will be notified. In such a situation where the lighting and shooting conditions have changed at two different times and the image is crowded, this model should be able to draw boundaries around the added or subtracted objects and introduce them to the user. If the presence of the object in the environment is necessary and the system correctly detects its removal or addition, informing the relevant expert about this safety issue is very helpful (for example, removing the fire capsule, adding an explosive, etc.). Alternatively, the system can be used to gauge the progress of an operational project. It means measuring what important changes have taken place in that environment at two different times. There are many such applications.

So, in general, we need a model that, by receiving the first and second images, identifies the objects that are in the first but not in the second, and vice versa. This model must be resistant to scene changes.

In this paper, a method based on deep learning with supervision is presented. It also explains the process of preparing natural and synthetic data sets and the process of labeling them. In the third part, the model is clearly explained and in the fourth part, the results are examined on various experimental data. The innovations of our work are:

- Expressing, paying attention to, and defining new problems in the form of finding the conceptual difference between the two images for use in spatial care projects and having the appropriate accuracy and resistance for use in different domains.

- Introduce the data collection process for this issue and also prepare the mentioned data. Also, introduce a process for data redundancy and synthesize it for use in high parameter models and increase the generalizability of the proposed models.

- Design and present a model based on neural networks for supervised learning and conversion of received images into 4 type change bitmap images.

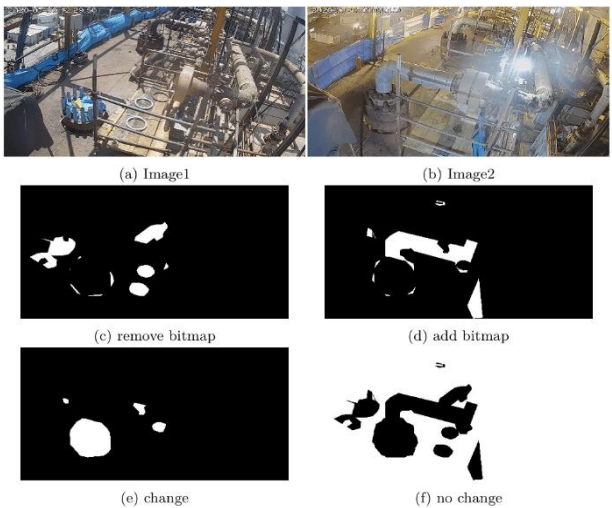

Fig. 1. The first two images are images of an environment at two different times. There are four black-and-white images whose pixels contain maps for adding, subtracting, changing, and not changing the scene, respectively.

## II. RELATED WORKS

In order to find the conceptual difference between the two scenes, the previous related methods in the field of detecting anomalies in the video or motion detection in the scene and the like can be used. The main need for this issue is the exact location of the different areas. Therefore, location methods, as well as segmentation methods, are carefully examined.

### A. Location by comparing patches

In the field of detecting and locating anomalies in the video, different tasks of comparing and modeling the scene behavior in different and fixed patches from different video frames have been studied [3, 4]. Under the influence of this type of method, for the problem of this article, the specifications of each patch from the second image can be compared with the specifications of the similar patch in the first image and their differences can be calculated. But these methods have a high location error.

### B. Location by segmentation

In these methods, which attribute each pixel of the scene to one of the components, the location accuracy is higher, but the implementation of these methods requires a lot of data from the range of images of the operating environment. Following this problem, different methods of unsupervised segmentation have been introduced [5, 6, 7], the use of which is the problem of this article can improve the results in the accuracy of locating differences.

### C. U-NET neural networks

By receiving an image, this structure first extracts the feature with a function consisting of several layers of neural network and makes the dimensions smaller than the input image or vector, then enters it into larger dimensions or its size and expects the desired output. An example of this structure can be seen in many self-encryption networks [8, 9, 10]. The benefits of this network structure are also evident in image segmentation. The advantage of these methods is in finding the connection of different pixels and structures within the image, thus reducing the location error [11, 12].

## III. PROPOSED METHOD

Images must be tagged to use supervised learning. For instance, when we have two images, if an object is removed from image one, it must be specified, and if an object is added to the second image, it must be specified. Also, if a change is made in one place (ie, one object is removed from the first and another object is added in the second), it must be specified as a change. For this purpose, a data set must be prepared in which these three items are specifically labeled.

Once the data set is provided, a model should be designed that takes the two images and then determines where the deleted object is removed from the first image and where the second image is added, and where a change has occurred from the two images.

### A. Making dataset

This section uses software that can label multifaceted objects. In this way, the two images No. 1 and 2 were first glued together and delivered to the users responsible for labeling. Users marked around the objects that are clearly present in the image on the left but not in the image on the right. Also, places where two images have a big conceptual difference, are considered with the change label.

Using the position of the labeled locations, four black&white images (bitmaps) with pixel values 0 and 1 were created. According to Fig.1 the first bitmap shows the location of the pixels containing the removed objects (objects in the first image but not in the second). the second map contains added objects (objects that are in the second but not in the first) and the third is a map of changes in two scenes (The pixels in the first image contain an object, which in the second image contains the same pixels as another object) and a fourth is a map of no changes (pixels in Both images did not change conceptually.

### B. Synthesize data

Due to the limited data and the difficulty and time of the labeling process to improve the model, the synthesis of data leads the model to the target more quickly.

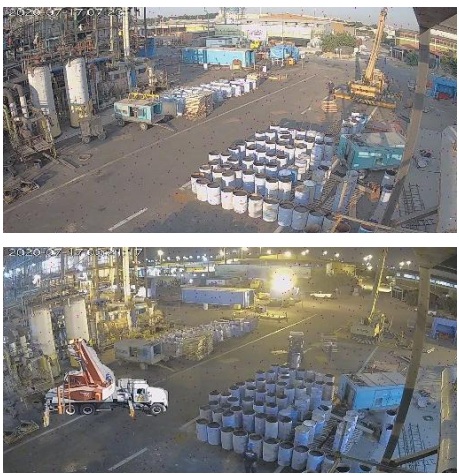

Fig. 2. An example of adding shapes by knowing where to add them to automatically generate the relevant bitmaps. In this image, a white truck has been added to the second image and just in the add bitmap the corresponding pixels have the value of 1.

First, a collection of objects can be added by providing a set of images of different objects and then adding noise and adjusting their darkness and light, as well as blurring their background.

Now by adding these objects to the first image, the pixels of the deleted map in that peer-to-peer area is filled with a value of 1, and by adding these objects to the second image, the map of the addition in the corresponding area is filled with a value of 1. If the result of the AND operator on them is 1, the map of the change in location will have the same value as 1. An example of this trend is shown in Figure 2.

In this way, The dataset can be expanded to any number, but it should be noted that the variety of objects added corresponds to the number of images, so that the model does not over-fit with the specific shape of these objects.

### C. Our method

In order to have a model that receives the two desired images and can produce the four black & white maps, considering that the color of the objects is very important, then all three color channels of the images must be used. We also need to resize the images to simplify them so that the details are not removed. The size is 256 by 512 for images.

As illustrated in Fig.3 the network input is considered as a set of 6-dimensional images, the first 3 dimensions of which are considered from the first image (image from the first time) and the next 3 dimensions from the second image (later image).

The final output of the model will be 4D. These dimensions include add, delete, change, and no change binary maps, If we consider **f** as a function to extract a feature from two source images and consider **g** as the function that produces the four maps of the above changes by receiving the output of the function **f** respectively.

$$g(f(x_1, x_2)) = (z_1, z_2, z_3, z_4)$$

Due to the transformation of the problem into supervised learning, we have the correct reference images as Y = (y1, y2, y3, y4). Therefore, by setting the parameters of g and f functions, we should try to minimize the error function.

The error function is defined by the following formula. Y consists of four change maps prepared manually or synthetically. The result of the g function is also defined as four bitmap maps. Therefore, the difference between these feature maps should be considered as a model error for training.

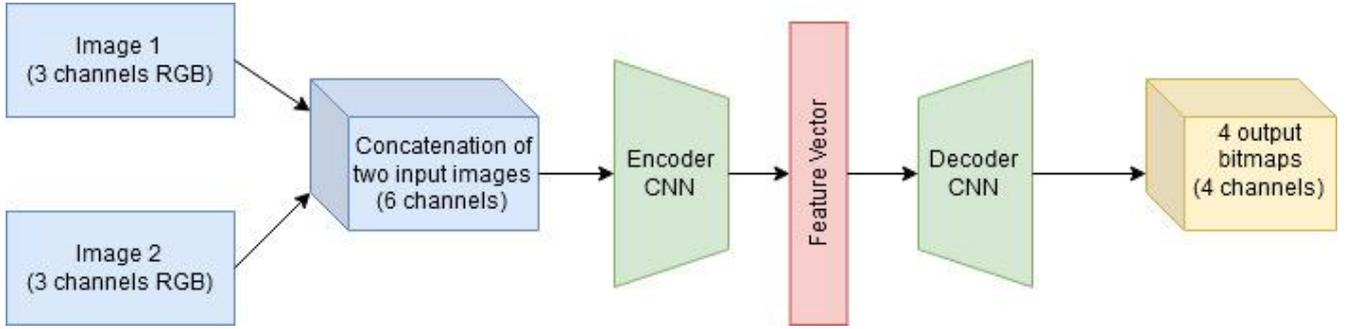

Fig. 3. Outline of the proposed method. Two different images belong to different times of a same location concatenate with eachother. Feature extractor network causes feature vestor and the decoder CNN creates four bitmaps in 4 channels of the output. 4 bitmaps are "ADDED","REMOVED","CHANGED","NOT CHANGED".

$$\min_{\theta_f, \theta_g} (Y - g(f(x_1, x_2)))^2$$

The model is designed as UNET. So that the function f is designed as a neural network with 6 convolutional blocks with the design of 16, 32, 64, 128, and 256 as encoder networks, and also g is designed as 3 convolutional blocks with the design of 256, 128, 64, 32, and 16 as decoder networks. As shown in Figure 3. The contents of each block include convolution layers, batch normalization, and convulsions with step 2.

## IV. EXPERIMENTS

In this section, 200 pairs of real and synthetic photographs are considered for testing. Also, precision, recall, and accuracy criteria were considered the most important criteria for comparing the results.

The results of different models and their configurations are expressed in different diagrams. According to the images obtained from the model on the test images, it is clear that an obvious improvement has been created by using these models.

Figure 4 shows a view of the model result on two sample images. As can be seen, objects that are significant in size are identified as added objects, if not in the first image, but in the second image, and vice versa in the first image as deleted objects.

More than 10 models with different UNET configurations were used and compared, which are presented separately with different images with precision-recall diagrams.

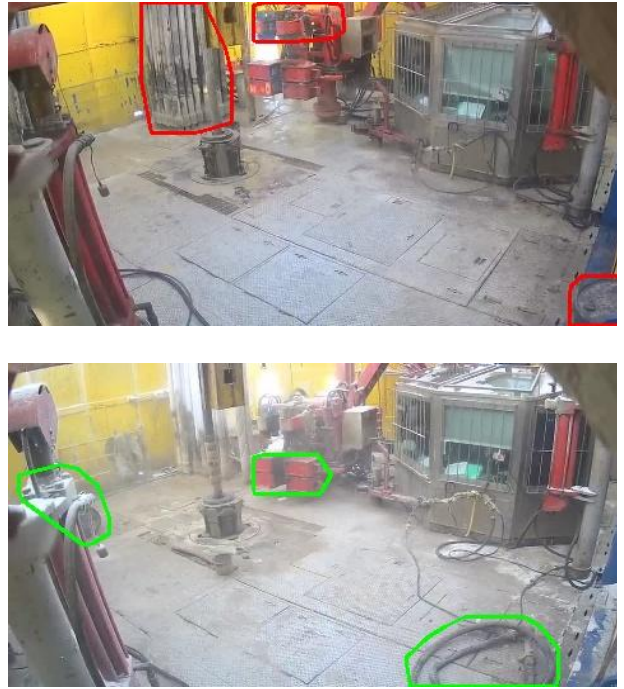

Fig. 4. The objects added in the image below and the deleted objects above were marked.

Due to the lack of a previous method for this project, the comparison results are presented as a comparison of different settings.

The effect of network depth on the perception of detail is determined by the diagrams. The greater the number of convolutional layers, the more detailed the model is. But due to the limitations of various data, the depth can not be exceeded.

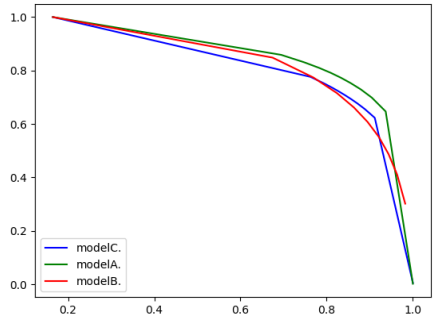

Fig. 5. Precision - recall curve for the map of added objects

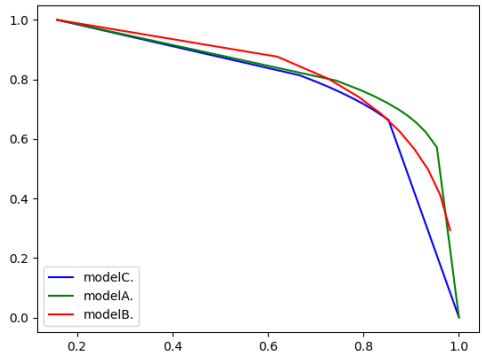

Fig. 6. Precision - recall curve for the map of removed objects

The effect of each of the input layers is such that in each output task the pixel value is specified. This reduces the need for the model to a large number of parameters.

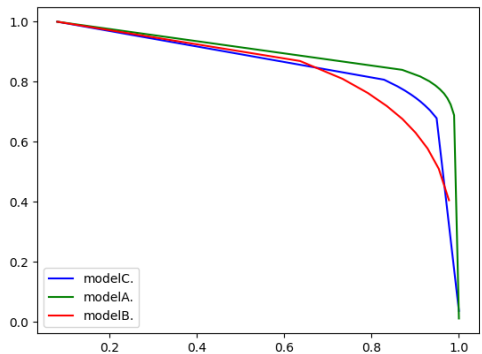

Fig. 7. Precision - recall curve for the map of changed objects

Because in the natural state, one of the additional modes of little change or no change can occur, which we have embedded in the map model for each. The impact of this can also be determined by evaluation.

Three different types of settings were examined in these comparisons, which are introduced as models A, B, and C. Its performance results are shown in Figures 5, 6, 7, and 8.

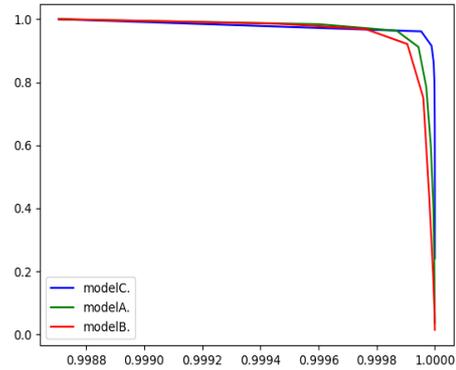

Fig. 8. Precision - recall curve for the map of not-changed objects

Model A consists of 6 levels of convolutional blocks used as 6,16,32,64,128,256 for the function f and 6 similarly symmetric blocks for g, respectively.

Model B consists of 4 convolutional blocks with a number of convulsions of 6, 16, 32, 64 for f and symmetric and similar for g, respectively.

Model C consists of 5 convolutional blocks 6,16,32,64,128 respectively Used for f and symmetric and similar for g.

According to the introduced diagrams, model A has the best performance for detecting addition, deletion, and change, but model C has shown better performance in detecting no change.

V. CONCLUSION

Observing the results, it is clear that with the limited depth of UNET settings, we have reached acceptable graphical results in all four maps. If we want to understand only the change, as can be seen from the diagram of the results related to the unchanged objects, acceptable results can be obtained.

Also, to improve the conceptual understanding of images, the use and adjustment of pre-trained feature extractors can be very helpful, but by modifying the error functions, the model should be limited to a few clusters of variable objects, which will be considered by future designers' work.